\documentclass[10pt,twocolumn,letterpaper]{article}

\usepackage[pagenumbers]{wacv} 

\usepackage{graphicx}
\usepackage{amsmath}
\usepackage{amssymb}
\usepackage{booktabs}

\usepackage[dvipsnames]{xcolor}
\usepackage{multirow}
\usepackage{colortbl}
\usepackage{comment}
\usepackage{times}
\usepackage{epsfig}
\usepackage{xcolor}
\usepackage{boldline}
\usepackage{algpseudocode}
\usepackage{algorithm}
\definecolor{wacvblue}{rgb}{0.21,0.49,0.74}
\usepackage[pagebackref,breaklinks,colorlinks,allcolors=wacvblue]{hyperref}

\title{PipeFlow: Pipelined Processing and Motion-Aware Frame Selection for Long-Form Video Editing}

\author{Mustafa Munir\\
The University of Texas at Austin\\
{\tt\small mmunir@utexas.edu}
\and
Md Mostafijur Rahman\\
The University of Texas at Austin\\
{\tt\small mostafijur.rahman@utexas.edu} \\
\and
Kartikeya Bhardwaj\\
Qualcomm AI Research\\
{\tt\small kbhardwa@qti.qualcomm.com} \\
\and
Paul Whatmough\\
Qualcomm AI Research\\
{\tt\small pwhatmou@qti.qualcomm.com} \\
\and
Radu Marculescu\\
The University of Texas at Austin\\
{\tt\small radum@utexas.edu} \\
}
\begin{document}
\maketitle

\begin{abstract}
Long-form video editing poses unique challenges due to the exponential increase in the computational cost from joint editing and Denoising Diffusion Implicit Models (DDIM) inversion across extended sequences. To address these limitations, we propose \textbf{PipeFlow}, a scalable, pipelined video editing method that introduces three key innovations: First, based on a motion analysis using Structural Similarity Index Measure (SSIM) and Optical Flow, we identify and propose to skip editing of frames with low motion. Second, we propose a pipelined task scheduling algorithm that splits a video into multiple segments and performs DDIM inversion and joint editing in parallel based on available GPU memory. Lastly, we leverage a neural network-based interpolation technique to smooth out the border frames between segments and interpolate the previously skipped frames. Our method uniquely scales to longer videos by dividing them into smaller segments, allowing PipeFlow's editing time to increase linearly with video length. In principle, this enables editing of infinitely long videos without the growing per-frame computational overhead encountered by other methods. PipeFlow achieves up to a 9.6$\times$ speedup compared to TokenFlow and a 31.7$\times$ speedup over Diffusion Motion Transfer (DMT).

\end{abstract}

\section{Introduction}
\label{sec:intro}

Generative artificial intelligence has recently witnessed substantial progress in image and video synthesis \cite{dmt, mahmud2024ada, liu2024sora, croitoru2022diffusion, pnpDiffusion2023, geyer2023tokenflow}, unlocking numerous applications in creative and professional fields such as content creation, real-time video editing \cite{streamv2v, kodaira2023streamdiffusion}, and style transfer \cite{StyleGAN}. In particular, latent diffusion models (LDMs) like Stable Diffusion \cite{Stable_Diffusion} have become leading methods for high-quality, realistic image generation \cite{pnpDiffusion2023, meng2022sdedit, saharia202_imagen}. These models are increasingly being adopted for video-based tasks, where the preservation of spatial and temporal coherence presents unique challenges.

\begin{figure}[t]
\centering
\includegraphics[width=1.05\columnwidth]{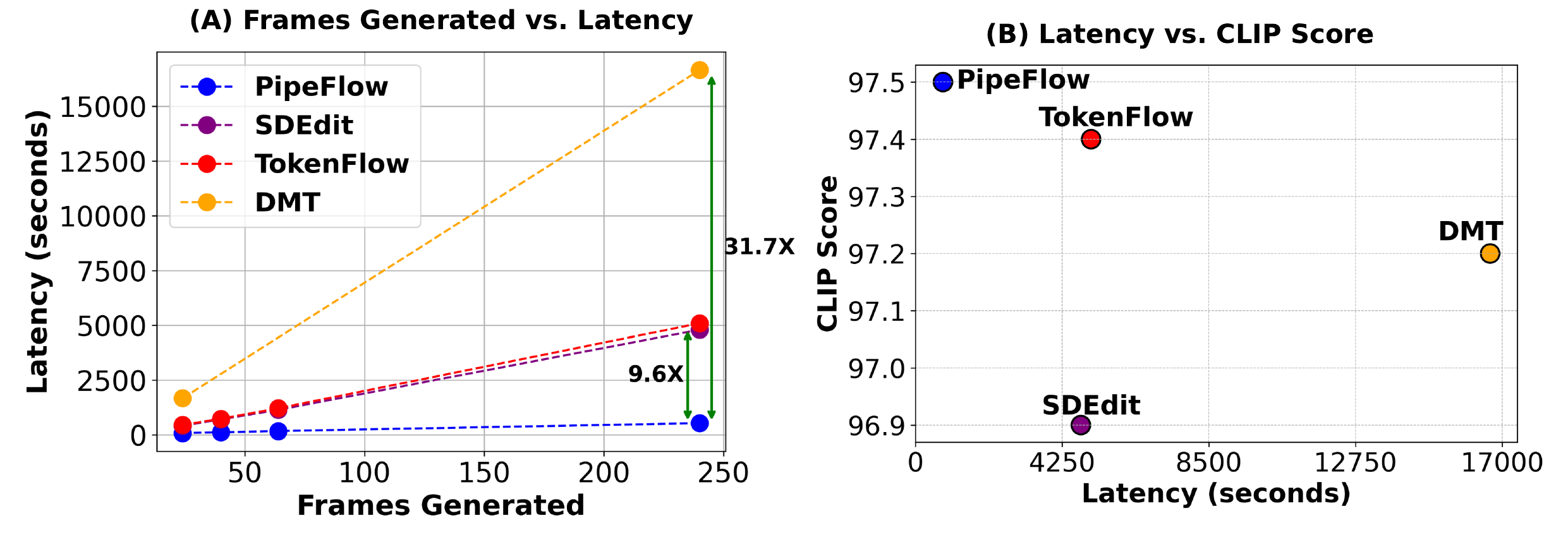}
\caption{\textbf{Single-GPU Latency Comparison.} \textbf{A:} Number of frames generated and the time needed to generate them for various methods. PipeFlow achieves the best performance compared to other state-of-the-art methods, achieving up to a 31.7$\times$\ speedup compared to DMT. \textbf{B:} Comparison of the time needed to generate 240 frames with the average CLIP score of generated frames. PipeFlow achieves the best latency to CLIP Score tradeoff.} 
\label{fig:pareto}
\vspace{-3mm}
\end{figure}

Despite these recent advances, existing diffusion-based approaches for video editing face critical limitations in terms of scalability and efficiency, especially for long-form and real-time applications \cite{geyer2023tokenflow, mahmud2024ada}. One key issue is the need for Denoising Diffusion Implicit Models (DDIM) inversion, which introduces a prohibitive computational cost for long video sequences \cite{pnpDiffusion2023}. Additionally, methods like TokenFlow \cite{geyer2023tokenflow} suffer from high memory consumption during joint editing of multiple frames, limiting their speed and usability for resource-constrained applications. 

Another challenge of long-form video editing is the need to maintain coherence and continuity across extended frames~\cite{liang2023flowvid, munir2025objectalign}. Consistent character representation and avoiding visual artifacts become increasingly challenging  as well in long-form video, often resulting in flickering or distortions~\cite{streamv2v, kodaira2023streamdiffusion, munir2025objectalign}. Moreover, the computational workload scales exponentially with video length~\cite{dmt}, necessitating efficient utilization of memory and processing power~\cite{geyer2023tokenflow}. 

The techniques proposed so far for consistent editing across an entire video often fail to scale effectively, as they struggle to balance video length and computational cost, leading to compromises in processing speed and quality \cite{geyer2023tokenflow, dmt}. Addressing these limitations is thus essential to enable efficient video editing.

To address these limitations, this paper introduces \textbf{PipeFlow}, a novel pipeline-based approach to enable efficient, temporally consistent, and scalable video editing with diffusion models. Indeed, by segmenting the frames for DDIM inversion and joint editing, our approach mitigates the exponential cost increase associated with these processes on longer videos. This extends the practical utility of diffusion-based editing for faster long-form video generation. We introduce three key innovations: 

\begin{itemize}
    \item First, we perform motion detection using Structural Similarity Index Measure (SSIM) and Optical Flow to identify frames with low motion, which are then skipped during editing. Skipped frames are later interpolated using a neural network to maintain smooth temporal continuity. 
    \item Second, we propose a pipelined, queue-based task scheduling algorithm that splits a video into \textit{N} segments, performs DDIM inversion and joint editing in parallel based on the available GPU memory. This approach allows PipeFlow to scale efficiently to long-form videos and multi-GPU setups, while avoiding the exponential computational costs of other state-of-the-art methods. 
    \item Lastly, we use neural network based frame interpolation between the boundaries of video segments and between the skipped low motion frames. This interpolation between adjacent segments helps us preserve visual coherence throughout the video.
\end{itemize}

\noindent
Overall, PipeFlow delivers a significant performance boost, achieving a 9.6$\times$ speedup over TokenFlow \cite{geyer2023tokenflow}, a 9.1$\times$ improvement compared to SDEdit \cite{meng2022sdedit}, and a 31.7$\times$ speedup compared to Diffusion Motion Transfer (DMT) \cite{dmt}, as illustrated in Figure~\ref{fig:pareto}A. PipeFlow also provides the best tradeoff between video quality and generation time, as shown in Figure~\ref{fig:pareto}B. When scaling from a single GPU to an 8-GPU setup, PipeFlow efficiently leverages multiple GPUs through segmentation and pipelined processing, achieving a  62.4$\times$ total speedup over TokenFlow \cite{geyer2023tokenflow}.

\vspace{-2mm}

\section{Related Work}
\label{sec:rel_work}

\subsection{Video Editing Methods}
 While Generative Adversarial Networks (GANs) initially dominated tasks such as super-resolution \cite{wang2018esrgan}, video enhancement \cite{jiang2021enlightengan}, and style transfer \cite{StyleGAN, wang2020imaginator, bar2022text2live, gupta2022rv}, Denoising Diffusion Probabilistic Models (DDPMs) \cite{ddpm} have emerged as the new state-of-the-art approach. DDPMs overcome traditional GAN limitations like training instability and mode collapse \cite{beatgan}, demonstrating remarkable progress in both image \cite{saharia202_imagen, pnpDiffusion2023} and video generation \cite{controlnet, bar2023multidiffusion, ho2022imagen_video}. Recent work has focused on improving frame consistency through innovations such as cross-frame attention \cite{geyer2023tokenflow} and optical flow integration \cite{liang2023flowvid}. However, efficiently scaling these methods to longer sequences remains a significant challenge \cite{geyer2023tokenflow, zhang2023controlvideo, wu2022tuneavideo, dmt, liang2023flowvid}.

Many video editing frameworks leverage pre-trained text-to-image generation models~\cite{pnpDiffusion2023, meng2022sdedit}, often based on Stable Diffusion~\cite{Stable_Diffusion}. Examples include Text2Video-Zero~\cite{text2video-zero, pnpDiffusion2023}, TokenFlow~\cite{geyer2023tokenflow}, StreamV2V~\cite{streamv2v}, Pix2Video~\cite{Ceylan2023Pix2VideoVE}, and Tune-A-Video~\cite{wu2022tuneavideo}. Two notable image editing methods, Plug and Play Diffusion (PNP)~\cite{pnpDiffusion2023} and SDEdit~\cite{meng2022sdedit}, serve as foundations for many training-free video editing models \cite{geyer2023tokenflow, mahmud2024ada}. PNP \cite{pnpDiffusion2023} provides a framework for text-guided image-to-image translation, adapting pre-trained diffusion models to maintain the semantic structure of a guidance image while following a target text prompt.

\vspace{-2mm}

\subsection{Pipelining \& Task Scheduling}
Pipelining is a fundamental technique in computer systems, widely utilized in modern CPUs and GPUs to facilitate parallel processing~\cite{instruction_parallelism, tomasulo1967efficient}. Task scheduling plays a crucial role in optimizing pipeline performance. Queue-based task scheduling is particularly effective for managing parallel processes and handling computationally intensive tasks~\cite{task_scheduling, queue_task_scheduling}. By organizing tasks into queues, the system decouples different pipeline stages, allowing for concurrent execution~\cite{queue_task_scheduling}. This dynamic resource allocation ensures efficient use of GPU memory, enabling the system to process multiple tasks as resources become available~\cite{task_scheduling}. Additionally, it minimizes bottlenecks, improving throughput and reducing delays~\cite{queue_task_scheduling, littles_law, orig_propose_little_law_style_idea, proof_littles_law}. 

In this paper, we propose a pipelined queue-based scheduling approach, segmenting videos and executing DDIM inversion and joint editing in parallel, based on memory availability. This approach alleviates computational bottlenecks, enabling efficient processing of long-form videos with minimal overhead.

\section{Preliminaries for Video Editing Techniques}
\label{sec:prelim}

\subsection{Denoising Diffusion Implicit Models (DDIM)}
DDPMs~\cite{nichol2021glide, ddpm, nichol2021improved} iteratively denoise an initial Gaussian noise image, $\mathbf{X}_T \sim \mathcal{N}(0, I)$, to reconstruct a target image, ${x}_0$ \cite{mahmud2024ada}. Diffusion methods have evolved with Latent Diffusion Models (LDMs)~\cite{Stable_Diffusion, blattmann2023videoldm} that enable generation in a compressed latent space, often using self-attention \cite{vaswani2017attention}, and text-guided cross-attention \cite{blattmann2023videoldm, mahmud2024ada}.

In the DDPM setup, image generation occurs over multiple timesteps \( T \). Denoising Diffusion Implicit Models (DDIMs)~\cite{song2020_ddim} provide a sampling method through both \textit{forward} and \textit{reverse} diffusion phases. In the forward phase, noise is incrementally added across \( T \) timesteps, while in the reverse phase, a model predicts and removes noise to reconstruct the original image. The computational cost of the reverse process depends on the model’s parameters \( O(N) \) and the number of timesteps \( T \), resulting in a complexity of \( O(T \cdot N) \) for each frame. Breaking down this cost further, if the model comprises \( n \) tokens per frame and \( d \) dimensions per token, the total DDIM inversion cost scales as \( O(T \cdot n^2 \cdot d) \cdot F\) for \( F \) frames. Joint editing involves additional computational layers, with self-attention adding a cost of \(O(n^2 d)\) per frame. While, cross-attention for \( K \) keyframes incurs a cost of \(
O(K \cdot n^2 d)\). With a token propagation cost of $O(n^2)$ we get a total cost of:

\vspace{-3mm}

\begin{equation}
\label{eq:Total_cost}
O(K n^2 d) + O(n^2) + O(T n^2 d) \cdot F\
\end{equation}

By turning the sequential process into a parallel process, our proposed PipeFlow is able to reduce the cost of Equation \ref{eq:Total_cost} to the cost of Equation \ref{eq:PipeFlow_Cost} as seen in Section \ref{subsec:pipelining}.

\begin{figure*}[ht]
\centering
\includegraphics[width=0.85\linewidth]{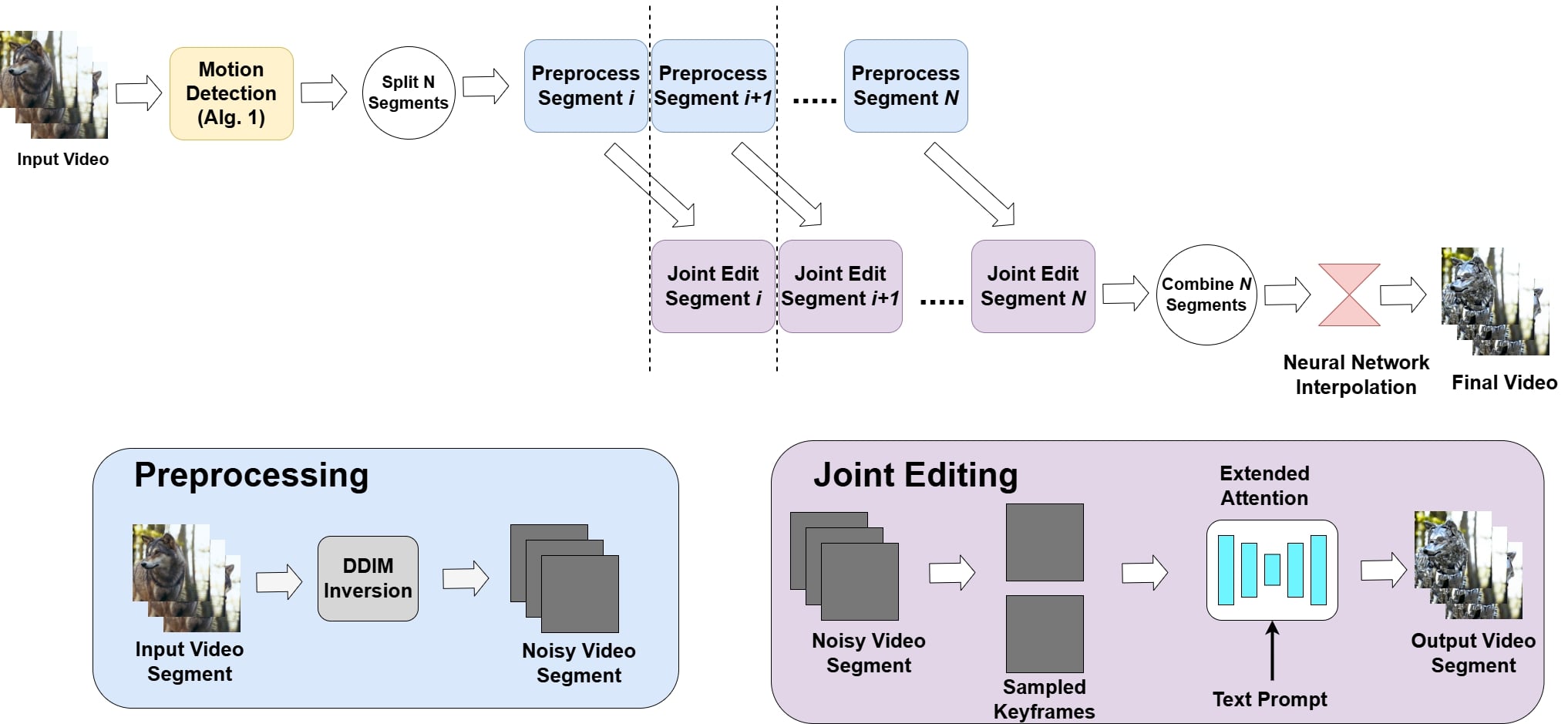}
\caption{\textbf{PipeFlow Overview.} The input video is analyzed for motion between frames (Section \ref{subsec:motion_detection}). Frames with low motion are skipped during editing and later interpolated. The video is then divided into \textit{N} segments for editing. Each segment undergoes DDIM inversion as preprocessing. Once the noisy video segments are created, the keyframes are sampled for joint editing. After the first preprocessing pass, joint editing for segment \textit{i} is performed in parallel with preprocessing for segment \textit{i}+1, continuing until all \textit{N} segments are processed.}
\label{fig:PipeFlow}
\vspace{-3mm}
\end{figure*}

\subsection{Extended Self-Attention}

Techniques such as SDEdit~\cite{meng2022sdedit}, ControlNet~\cite{controlnet}, and PNP~\cite{pnpDiffusion2023} offer controlled image editing, but fall short when applied to video editing due to their inability to maintain temporal consistency between frames. To address this issue, extended self-attention (ESA)~\cite{geyer2023tokenflow, peng2024conditionvideo, mahmud2024ada} introduces a method for improving temporal stability by extending attention across frames. ESA can be expressed as follows:

\vspace{-3mm}

\begin{align}
& \mathtt{ESA}(X_i) = \mathtt{\Omega} \left(\frac{Q_i K_{\mathtt{all}}^T}{\sqrt{d}}\right) V_{\mathtt{all}}, \ \ \ \forall i = \{1, \dots, N \} \\
\nonumber
\end{align}

\vspace{-3mm}

where, \(Q^{(i)}\), \(K^{(i)}\), and \(V^{(i)}\) correspond to the query, key, and value matrices of the \(i\)-th frame. \(K^{(\text{all})}\) and \(V^{(\text{all})}\) are the keys and values aggregated across all frames, allowing the model to maintain temporal consistency by incorporating information from the entire video sequence. Lastly, \(\mathtt{\Omega}(z)\) represents the softmax \cite{softmax} operator to ensure that attention weights are properly normalized.

However, cross-frame attention also incurs high memory and computational costs due to extending keys and values across all frames, especially for longer videos. To reduce this cost, our approach focuses on balancing the joint editing and DDIM inversion process through a pipelined approach, simultaneously decreasing the cost of both.

\section{Methodology}
\label{sec:methods}

\subsection{Motion Detection in Video Streams}
\label{subsec:motion_detection}
\paragraph{Motion-Aware Frame Selection}

To efficiently process long-form videos, we introduce a \textit{motion-aware frame selection} mechanism that identifies and allows us to skip frames with minimal visual changes, significantly reducing the computational overhead, yet without compromising the video quality. Our approach combines two complementary metrics: SSIM \cite{ssim} and Optical Flow magnitude \cite{farneback2003two}, providing robust motion detection across various scenarios.

\vspace{-4mm}

\paragraph{Frame Analysis Pipeline}
Given two consecutive frames $F_t$ and $F_{t+1}$ (at times $t$ and $t+1$), we first convert them to grayscale using the standard luminance conversion:

\begin{equation}
    G_t = RGB2Gray(R_t, G_t, B_t)
\end{equation}

where $R_t$, $G_t$, and $B_t$ represent the red, green, and blue channels, respectively, at time $t$.

\vspace{-4mm}

\paragraph{Structural Similarity Analysis}
We use SSIM to measure the structural similarity between consecutive frames. For two grayscale frames $G_t$ and $G_{t+1}$, SSIM is computed as:

\begin{equation}
    \footnotesize
    SSIM(G_t, G_{t+1}) = \frac{(2\mu_t\mu_{t+1} + C_1)(2\sigma_{t,t+1} + C_2)}{(\mu_t^2 + \mu_{t+1}^2 + C_1)(\sigma_t^2 + \sigma_{t+1}^2 + C_2)}
\end{equation}

where:
\begin{itemize}
    \item $\mu_t, \mu_{t+1}$ are the mean intensities of frames $G_t$ and $G_{t+1}$.
    \item $\sigma_t^2, \sigma_{t+1}^2$ are the variances of frames $G_t$ and $G_{t+1}$.
    \item $\sigma_{t,t+1}$ is the covariance between frames $G_t$ and $G_{t+1}$.
    \item $C_1, C_2$ are constants to prevent division by zero.
\end{itemize}

\vspace{-4mm}

\paragraph{Optical Flow Analysis}
We complement SSIM with optical flow analysis using the Farneback algorithm to capture motion dynamics \cite{farneback2003two}. The Farneback algorithm computes a dense optical flow field $\mathbf{F}(x,y)$ between two consecutive frames, where each pixel $(x,y)$ has a displacement vector $\mathbf{F}(x,y) = (u_{x,y}, v_{x,y})$. The \textit{mean flow magnitude} $M_F$ is calculated as:

\begin{equation}
    M_F = \frac{1}{WH}\sum_{x=1}^W\sum_{y=1}^H\sqrt{u_{x,y}^2 + v_{x,y}^2}
\end{equation}

where $W$ and $H$ are the frame width and height, respectively, and $u_{x,y}$, $v_{x,y}$ denote the horizontal and vertical components of the flow vector $\mathbf{F}(x,y)$.

\vspace{-4mm}

\paragraph{Adaptive Motion Detection}
Given thresholds $\tau_s$ for SSIM and $\tau_f$ for optical flow magnitude, a frame is considered to have significant motion if either of the following conditions is met:

\vspace{-4mm}

\begin{equation}
    \text{Motion Detected} = \begin{cases}
        1 & \text{if } SSIM(G_t, G_{t+1}) < \tau_s  \\
        1 & \text{if } M_F > \tau_f \\
        0 & \text{otherwise}
    \end{cases}
\end{equation}

\vspace{-4mm}

\paragraph{Frame Selection Strategy}
Our frame selection process, as shown in Algorithm \ref{alg:frame_selection}, identifies key frames by analyzing motion and structural similarity. Starting with an initial frame ($F_0$), the algorithm iterates through a sequence of frames $\{F_1, \dots, F_n\}$. Each frame $F_t$ is converted to grayscale ($G_t$), then the SSIM between the grayscale frames $G_{t-1}$ and $G_t$ measures structural changes. 

Simultaneously, the optical flow between these frames is computed. The mean flow magnitude ($M_F$) quantifies overall motion intensity. A frame is selected if its SSIM falls below ($\tau_s$), its $M_F$ exceeds ($\tau_f$), or it is the final frame ($F_n$). Lastly, the algorithm returns a set of indices $S$ corresponding to the selected frames.

\vspace{-2mm}

\begin{algorithm}
\caption{Motion-Aware Frame Selection}
\label{alg:frame_selection}
\begin{algorithmic}[1]
\Require Initial frame $F_0$, frame sequence $\{F_1, ..., F_n\}$, thresholds $\tau_s, \tau_f$
\Ensure Selected frame indices $S$
\State $S \gets \{0\}$ \Comment{Always include first frame}
\For{$t \gets 1$ to $n$}
    \State $G_t \gets \text{RGBToGray}(F_t)$
    \State $G_{t-1} \gets \text{RGBToGray}(F_{t-1})$
    \State $ssim \gets \text{SSIM}(G_{t-1}, G_t)$
    \State $flow \gets \text{OpticalFlow}(G_{t-1}, G_t)$
    \State $M_F \gets \text{MeanMagnitude}(flow)$
    \If{$ssim < \tau_s$ \textbf{or} $M_F > \tau_f$ \textbf{or} $t = n$} 
        \State $S \gets S \cup \{t\}$ \Comment{motion detected or final frame}
    \EndIf
\EndFor
\Return $S$
\end{algorithmic}
\end{algorithm}

\vspace{-3mm}

Our approach reduces the number of frames undergoing DDIM inversion and joint editing, thus decreasing the computational cost while maintaining temporal coherence.

\vspace{-1mm}
\subsection{Pipelining DDIM Inversion and Editing}
\label{subsec:pipelining}
Our approach implements an efficient pipelining strategy that accounts for the sequential dependency between DDIM inversion and joint editing. For each video segment, DDIM inversion must precede joint editing, creating an opportunity for parallel processing across segments. After DDIM inversion of the first segment, the segment undergoes joint editing; simultaneously, the DDIM inversion can begin on the second segment as shown in Figure \ref{fig:PipeFlow}. Depending on available GPU memory, we can parallelize additional tasks, such as performing DDIM inversion on multiple segments simultaneously. However, the fundamental constraint remains: a segment's DDIM inversion must complete \textit{before} its joint editing phase can begin.

The pipeline operates through overlapping execution phases that minimize idle time and maximize resource utilization. Starting with DDIM inversion of segment $i$, once completed, joint editing of segment $i$ commences simultaneously with DDIM inversion of segment $i+1$ as shown in Figure \ref{fig:PipeFlow}. This pattern continues throughout the video sequence, maintaining the ordering between segment while enabling parallel processing across segments.

The computational cost associated with our methodology is notably lower than in previous joint editing based frameworks \cite{geyer2023tokenflow, mahmud2024ada}. More precisely, the total cost can be expressed as: \(\text{Total cost} = O(T \cdot N)\), where \( N \) can be decomposed into \( n \) tokens per frame and \( d \) dimensions of tokens: \(\text{Total cost} = O(T \cdot n \cdot d)\). After considering the cost of self-attention, the total cost becomes: \(\text{Total cost} = O(T \cdot n^2 \cdot d)\). Furthermore, the costs associated with joint editing components are as follows:
\begin{itemize}
\item The self-attention mechanism cost is $O(n^2 d)$ per frame.
\item The cross-attention operation, on \( B \) keyframes in the batch, results in a cost of $O(B \cdot n^2 d)$.
\item The token propagation process adds a cost of $O(n^2)$.
\end{itemize}
Combining these costs, the \textit{overall computational cost} is:

\vspace{-2mm}

\begin{equation}
\label{eq:PipeFlow_Cost}
\small
O(T \cdot n^2 \cdot d)_i \rightarrow O(B \cdot n^2 d) + O(n^2)_i \quad \text{OR} \quad O(T \cdot n^2 \cdot d)_{i+1}
\end{equation}

where the subscript indicates the index of the segment/batch, the arrow indicates that the process on the right side must occur \textit{after} the process on the left side, and the \textit{OR} denotes parallel execution of the processes (where the cost becomes the lesser of the two since the next task can begin after either completes). 

The computational cost of our approach, as shown in Equation \ref{eq:PipeFlow_Cost}, is lower than the non-pipelined cost, which is presented in Equation \ref{eq:Total_cost}. This improvement is achieved because, after performing DDIM inversion on the first batch, our method enables both DDIM inversion and joint editing of subsequent batches to occur in parallel. By adopting this pipelined approach, we achieve not only a reduction in the total computational cost, but also an increase in the processing speed, facilitating the efficient handling of long-form video editing tasks. 

\vspace{-4mm}
\paragraph{Natural Multi-GPU Parallelism for Video Editing.} Additionally, segmenting videos into smaller batches enables PipeFlow to efficiently leverage multiple GPUs, unlike prior methods \cite{geyer2023tokenflow, dmt}. Previous frameworks \cite{geyer2023tokenflow, dmt} cannot efficiently leverage multiple GPUs because tensor parallelism is inefficient for diffusion models due to significant activation sizes and high communication overheads \cite{li2024distrifusion}. In contrast, our pipeline-based parallelism minimizes communication and distributes independent segments across GPUs, thus significantly enhancing multi-GPU efficiency.

\subsection{Queue-Based Task Scheduling}
\label{subsec:task_scheduling}

To optimize our parallel processing of DDIM inversion and joint editing, we propose a queue-based task scheduling approach. This methodology enhances our system's efficiency by facilitating asynchronous processing, allowing multiple DDIM inversions to occur simultaneously when joint editing resources are not currently available. Since joint editing relies on frames that have completed the DDIM inversion, this queuing mechanism ensures that we maintain a steady flow of tasks without idle processing time.

In our implementation, tasks are managed through a dynamic queue system that monitors the GPU memory usage during execution. Before initiating a new DDIM inversion, the system checks the available GPU memory to ensure that sufficient resources are allocated for the operation. If memory is constrained, the task is queued until the necessary resources become available. This process not only optimizes resource utilization, but also prevents overloading the GPU, which can lead to performance degradation.

Applying Little's Law from queueing theory \cite{littles_law}, we can express the relationship between the average number of tasks in the system (\(L\)), the average arrival rate of tasks (\(\lambda\)), and the average time a task spends in the system (\(W\)): \(L = \lambda W\). In our context, \(L\) represents the average number of tasks in the queue (both DDIM inversions and joint editing), \(\lambda\) is the rate at which new tasks (DDIM inversions) arrive, and \(W\) is the average time each task spends in the system. By employing an efficient queuing strategy, we can increase \(\lambda\) through concurrent DDIM inversions, thus reducing \(W\) as tasks are processed more quickly.

The total time for serial processing of tasks is given by:

\begin{equation}
T_{\text{serial}} = N_1 N_2 T_1 T_2,
\end{equation}

where \(N_1\) and \(N_2\) are the number of tasks involved in DDIM inversion and joint editing, respectively, and \(T_1\) and \(T_2\) are the respective processing times for each task. In contrast, the asynchronous processing time is as follows:

\begin{equation}
T_{\text{async}} = \frac{N_1 N_2 T_1 T_2}{B},
\end{equation}

where \(B\) is the number of batches the tasks are split into. The comparison of serial and asynchronous processing illustrates that by utilizing our queuing approach and asynchronous processing, we can significantly reduce the overall processing time for long-form video editing by allowing tasks to occur as soon as there is available GPU memory.

\begin{algorithm}[H]
\caption{Pipelined DDIM Inversion and Joint Editing with Queue-Based Task Scheduling}
\label{alg:pipelined_editing}
\small
\begin{algorithmic}[1]
\State \textbf{Input:} Video segments: \texttt{N\_list}, maximum jobs: \texttt{MJ}, minimum GPU memory threshold: \texttt{MEM}
\State Initialize \texttt{MJ}, \texttt{MEM}, \texttt{N\_list}
\State Initialize empty queue \texttt{task\_queue}

\For{\texttt{N} in \texttt{N\_list}}
    \If{\texttt{N+1} $\leq$ len(\texttt{N\_list})}
        \State Add DDIM for \texttt{N+1} to \texttt{task\_queue}
    \EndIf
    \State Add joint editing for \texttt{N} to \texttt{task\_queue}
        \If{\texttt{free\_mem} $\geq$ \texttt{MEM} \textbf{and} \texttt{running\_jobs} $<$ \texttt{MJ}}
            \State Select next task from \texttt{task\_queue} based on priority (DDIM for \texttt{N} or joint editing for \texttt{N-1})
            \If{DDIM task selected}
                \State Run DDIM on \texttt{N}

            \ElsIf{joint editing task selected}
                \State Run joint editing on \texttt{N-1}
            \EndIf
        \EndIf
\EndFor
\State \textbf{Wait} for all background jobs to complete
\end{algorithmic}
\end{algorithm}

Algorithm \ref{alg:pipelined_editing} shows how we utilize pipelining and task scheduling to handle a greater volume of tasks concurrently resulting in enhanced performance and efficiency. Specifically, our approach allows for up to \(MJ\) jobs to run in parallel, as long as the available GPU memory meets or exceeds the minimum required threshold (\(MEM\)).

\subsection{Neural Network Based Frame Interpolation}
\label{subsec:frame_interpolation}

\paragraph{Border Frame Interpolation}
When processing video segments independently, discontinuities can emerge at segment boundaries. To address this, we apply RIFE \cite{huang2022rife} to interpolate transition frames between segments.

Given two adjacent segments $S_i$ and $S_{i+1}$, we interpolate between the last frame of $S_i$ and the first frame of $S_{i+1}$:

\vspace{-1mm}

\begin{equation}
    F_{interp} = \text{RIFE}(S_i^{last}, S_{i+1}^{first})
\end{equation}

where $F_{interp}$ represents the interpolated transition frame(s). This smooths the transition between segments, reducing potential artifacts and maintaining visual coherence.

\begin{figure*}[ht]
\centering
\includegraphics[width=0.84\textwidth]{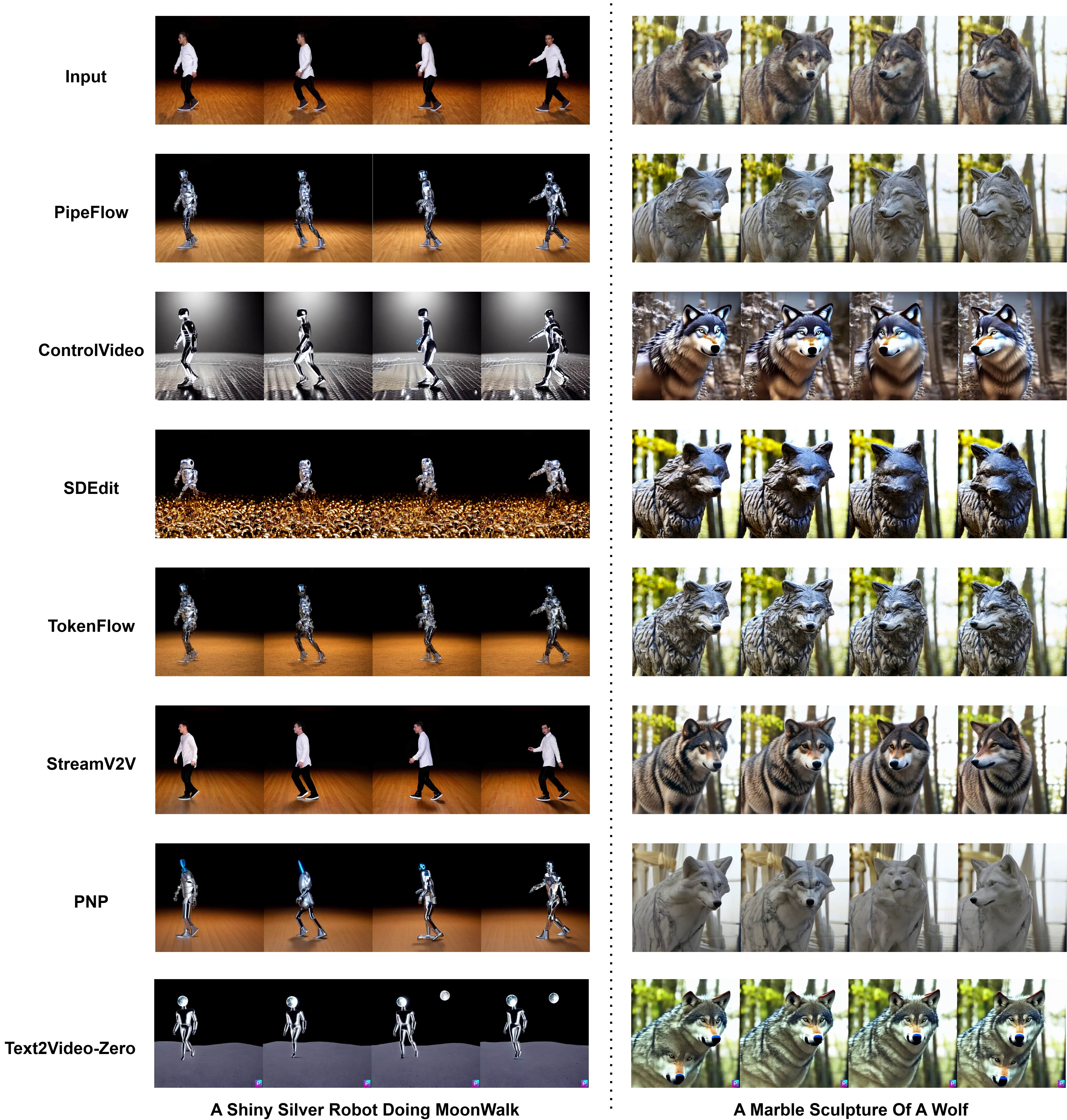}
\caption{\textbf{Qualitative comparisons} against competing state-of-the-art video editing methods. As shown, ControlVideo \cite{zhang2023controlvideo} suffers from inconsistent character generation. TokenFlow \cite{geyer2023tokenflow} shows better temporal consistency, but suffers from weaker edits not corresponding to the prompt as well. StreamV2V \cite{streamv2v} fails to follow the prompt properly and also suffers from larger amounts of flickering and motion blurring. Text2Video-Zero \cite{text2video-zero} is unable to edit longer form video and also fails to properly maintain the motion of the original video. \textbf{PipeFlow (ours)} achieves the best visual quality while maintaining consistency throughout the videos.}
\label{fig:Competing_Methods_Comp}
\end{figure*}

\vspace{-2mm}

\paragraph{Low-Motion Frame Recovery}
For frames skipped during the motion detection phase (Section \ref{subsec:motion_detection}), we utilize RIFE \cite{huang2022rife} to reconstruct the intermediate frames. Given two consecutive retained frames $F_t$ and $F_{t+k}$ where $k-1$ frames were skipped, we generate the intermediate frames through recursive interpolation:

\vspace{-2mm}

\begin{equation}
    \{F_{t+1}, ..., F_{t+k-1}\} = \text{RIFE}_{\text{recursive}}(F_t, F_{t+k}, k-1)
\end{equation}

The recursive interpolation process continues until all skipped frames are reconstructed, ensuring a smooth motion between the retained keyframes.

This dual application of RIFE \cite{huang2022rife} enables our pipeline to maintain temporal consistency while benefiting from the computational savings of our motion-aware frame selection strategy and our pipelined processing strategy.

\section{Experimental Results}
\label{sec:results}

We describe our experimental setup in Section \ref{subsec:setup}, our qualitative and quantitative results in Section \ref{subsec:qual_results} and Section \ref{subsec:quan_results}, and our ablation studies in Section \ref{subsec:ablations}.

\subsection{Experimental Setup}\label{subsec:setup}

We have implemented our method on top of PNP \cite{pnpDiffusion2023} as our image-to-image editing method. For our experimental evaluation, we use a total of 75 text-video prompt pairs videos from two primary sources: DAVIS \cite{davis} and Pexels \cite{pexels}. Specifically, the Pexels videos are obtained from the ReRender repository \cite{rerender}. The prompts we use draw inspiration from previous works such as \cite{mahmud2024ada, streamv2v, geyer2023tokenflow, text2video-zero, dmt, rerender}. The videos cover a wide variety of environments and subjects including animals and humans. The prompts cover a variety of edits including different stylizations and different object edits. We compare our method to other state-of-the-art video editing methods utilizing their default settings.

We note that our approach is generalizable and can be built on top of multiple methods such as TokenFlow \cite{geyer2023tokenflow} and Ada-VE \cite{mahmud2024ada}. Our approach can also be used with different versions of Stable Diffusion \cite{Stable_Diffusion} as PNP \cite{pnpDiffusion2023} can use many different stable diffusion model versions (e.g., Stable Diffusion v1.5, v2.0, v2.1, etc.). Our methods could also be used for speeding up and improving quality of other methods not based on PNP \cite{pnpDiffusion2023}, such as SDEdit \cite{meng2022sdedit}. Note that only 16 frames of generated videos were considered for Text2Video-zero \cite{text2video-zero} and 24 frames for DMT \cite{dmt} due to extensive memory requirements for Text2Video-Zero to operate on longer videos and because DMT \cite{dmt} states it only works on a maximum of 24 frames. Moreover, we considered TokenFlow integrated SDEdit \cite{meng2022sdedit} to make it a stronger video baseline for comparison.

\subsection{Qualitative Results}\label{subsec:qual_results}

We present a qualitative analysis of our method versus competing state-of-the-art methods in Figure \ref{fig:Competing_Methods_Comp}. In the example of the wolf, ControlVideo does not properly follow the prompt of making the wolf a marble sculpture and also generates inter-frame inconsistencies. SDEdit~\cite{meng2022sdedit} and TokenFlow~\cite{geyer2023tokenflow} often produce blurry hands in the moonwalk video. StreamV2V \cite{streamv2v} fails to follow the prompt in both videos and has significant flickering and temporal inconsistencies. Text2Video-Zero \cite{text2video-zero} has temporal inconsistencies (introduces a moon that is in some frames but not others), fails to follow the prompt (does not make the wolf marble, changes the background of the moonwalk video to outer space on the moon), and is unable to process longer videos.

In contrast, PipeFlow better follows the prompt of creating a marble sculpture of a wolf and has reduced flickering compared to other methods. These results demonstrate the superior quality of PipeFlow compared to other state-of-the-art methods.

\subsection{Quantitative Results}\label{subsec:quan_results}
\paragraph{Runtime:}

We benchmark the runtime of all methods with 512 $\times$ 512 resolution videos containing between 16 and 350 frames at an FPS of 20 on an Nvidia GeForce RTX 3090 GPU. As shown in Figures \ref{fig:pareto}A and \ref{fig:pareto}B, PipeFlow demonstrates significant performance improvements over existing methods. For shorter videos (24 frames), PipeFlow achieves 6.0$\times$ and 22.5$\times$ speedups compared to TokenFlow and DMT, respectively. 

These performance gains become even more pronounced with longer videos, where at 240 frames, PipeFlow achieves a 9.6$\times$ speedup over TokenFlow and a 31.7$\times$ speedup over DMT. This increasing performance gap demonstrates how our pipeline-based approach effectively addresses the scalability challenges that existing methods face with longer videos \cite{geyer2023tokenflow, mahmud2024ada, dmt}.

While TokenFlow and DMT show increasing computational overhead per frame as video length grows, PipeFlow maintains consistent per-frame processing times, requiring only 526 seconds to process 240 frames compared to TokenFlow's \cite{geyer2023tokenflow} 5091 seconds and DMT's \cite{dmt} 16650 seconds for the exact same task. DMT \cite{dmt} can operate on a maximum of 24 frames, so we report results for 24 frames and approximate its processing  times for longer videos through sequentially processing videos in groups of 24 frames. Unlike prior methods \cite{geyer2023tokenflow, text2video-zero, dmt} PipeFlow scales to arbitrary (infinite) length videos without any slowdown or out-of-memory (OOM) errors as videos can be split into segments that fit within GPU memory.

\noindent
\textbf{Multi-GPU Benefits of PipeFlow:} Existing video editing methods \cite{geyer2023tokenflow, dmt} struggle to efficiently leverage multi-GPU setups because tensor parallelism introduces significant communication overheads due to large activation sizes \cite{li2024distrifusion}. In contrast, PipeFlow’s pipelining allows each GPU to operate independently without communication overhead. Using 4 RTX A6000 Ada GPUs, PipeFlow achieves a \textbf{3.9$\times$} speedup, while TokenFlow \cite{geyer2023tokenflow}, lacking segmentation and pipelining, achieves only \textbf{1.1$\times$}. When combined with our existing \textbf{9.6$\times$} speedup compared to TokenFlow \cite{geyer2023tokenflow}, PipeFlow obtains a total \textbf{34.0$\times$} speedup over TokenFlow. These efficiency gains further scale with additional GPUs: On 8 GPUs, PipeFlow achieves a \textbf{7.8$\times$} multi-GPU speedup, significantly outperforming the \textbf{1.2$\times$} speedup of both TokenFlow \cite{geyer2023tokenflow} and DMT \cite{dmt}. Consequently, PipeFlow achieves total speedups of \textbf{39$\times$} and \textbf{62.4$\times$} over TokenFlow for short-form (24 frames) and long-form (240 frames) videos, respectively. Likewise, PipeFlow attains \textbf{146.3$\times$} and \textbf{206$\times$} speedups over DMT for short-form and long-form videos.

\begin{figure}[ht]
\centering
\includegraphics[width=1.0\columnwidth]{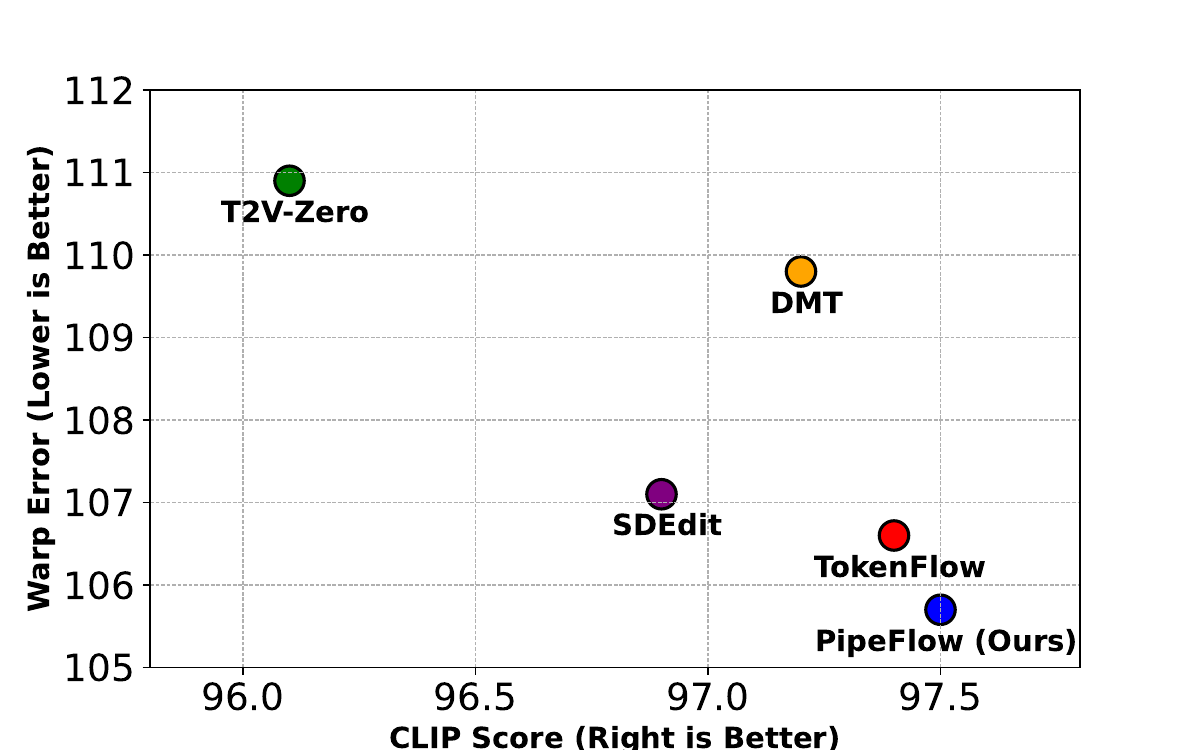}
\caption{\textbf{CLIP Score Vs. Warp Error Comparison}. We present a comparison of the CLIP Score and Warp Error of PipeFlow vs. competing state-of-the-art methods. Our PipeFlow achieves the best CLIP Score and Warp Error against all competing methods.}
\label{fig:CLIP_Warp}
\end{figure}

\noindent
\textbf{CLIP Score:} To evaluate the effectiveness of video editing methods in adhering to textual prompts and maintaining visual consistency, we use the CLIP score \cite{CLIP_Score}, following prior work \cite{streamv2v, mahmud2024ada}. This metric calculates the mean cosine similarity of CLIP image embeddings across all video frames, providing a measure of prompt alignment and temporal coherence. As shown in Figure \ref{fig:CLIP_Warp}, PipeFlow achieves the highest CLIP score of 97.5, outperforming TokenFlow \cite{geyer2023tokenflow} (97.4), DMT \cite{dmt} (97.2), SDEdit \cite{meng2022sdedit} (96.9), and Text2Video-Zero (96.1) \cite{text2video-zero}, demonstrating its superior capacity to produce consistent, high-quality edits.

\vspace{2mm}

\noindent
\textbf{Warp Error:}  
Following prior work \cite{geyer2023tokenflow, streamv2v, mahmud2024ada}, temporal consistency is evaluated using the warp error \cite{Warp_Error}. This metric quantifies the pixel-wise discrepancies between original and edited frames. Using a pre-trained optical flow estimator \cite{raft}, we compute the optical flow between consecutive frames in the original video and warp the corresponding edited frames. The warp error is then calculated as the average mean squared pixel error between the frames. As shown in Figure \ref{fig:CLIP_Warp}, PipeFlow achieves a warp error of 105.7, beating TokenFlow \cite{geyer2023tokenflow} (106.6), DMT \cite{dmt} (109.8), SDEdit \cite{meng2022sdedit} (107.1), and Text2Video-Zero (110.9) \cite{text2video-zero}.

\vspace{2mm}

\noindent
\textbf{Pareto Optimal Performance:}  
By achieving the best performance in both CLIP score and warp error while significantly reducing runtime, PipeFlow achieves a Pareto-optimal balance between speed, editing quality, and temporal consistency.

\begin{figure}[ht]
\centering
\includegraphics[width=\columnwidth]{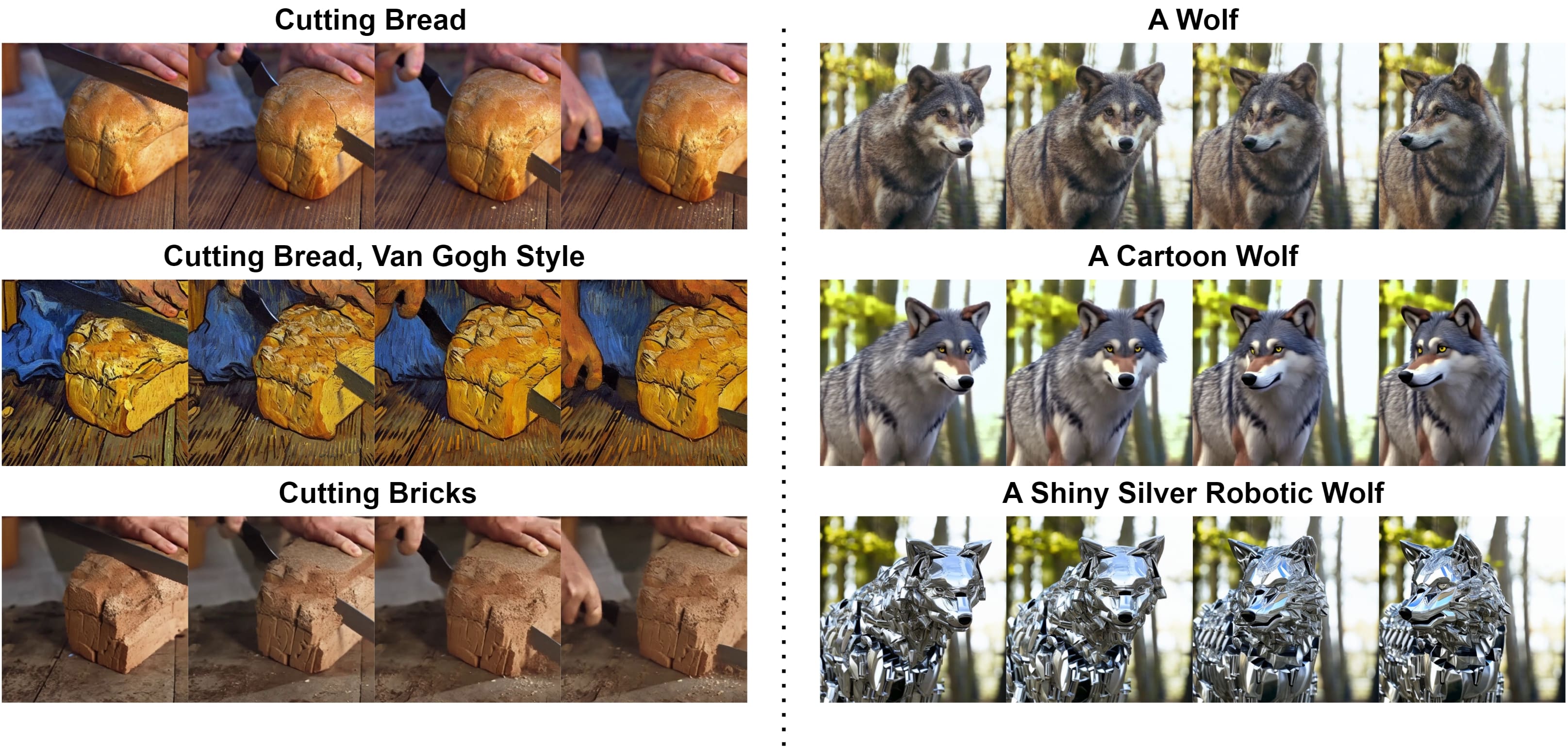}
\caption{\textbf{Ablation Study.} Ablation study on object editing and style transfer prompts. PipeFlow achieves strong visual quality, while maintaining consistency throughout the videos.}
\label{fig:ablation_video_edits}
\end{figure}

\subsection{Ablation Studies}\label{subsec:ablations}

\paragraph{Qualitative Analysis:} 
Figure~\ref{fig:ablation_video_edits} showcases a qualitative ablation study on various object editing and style transfer prompts. We evaluate both style transfer prompts, such as transforming videos into a cartoon or Van Gogh-inspired painting style, and object-editing prompts, such as altering a video of cutting bread to cutting bricks, and transforming a wolf into a  shiny silver robotic version. 

PipeFlow consistently applies the specified edits to the target objects and is capable of transforming the video style while preserving the fine-grain details of the moving subjects. The examples show that PipeFlow effectively modifies the subject in accordance with the prompt and maintains visual coherence throughout the video. This highlights PipeFlow's ability to produce aesthetically pleasing and detailed results, even with complex edits.

\section{Conclusion}
\label{sec:conc}

In this paper, we have addressed the computational challenges of long-form video editing by introducing \textbf{PipeFlow}, a scalable and efficient pipeline-based framework. Unlike previous methods that struggle with the exponential costs of sequentially performing DDIM inversion and joint editing, PipeFlow leverages three core innovations: motion-aware frame skipping, pipelined queue-based task scheduling, and neural network-based frame interpolation in between the pipelined segments. Our contributions enable significant computational savings while maintaining temporal consistency, thus allowing PipeFlow to provide high-quality edits even as video length increases.

{
    \small
    \bibliographystyle{ieeenat_fullname}
    \bibliography{main}
}

\clearpage
\setcounter{page}{1}
\maketitlesupplementary

\section{User Study}
\label{sec:user_study}

We conduct a user preference study to compare our method with competing state-of-the-art methods: TokenFlow\cite{geyer2023tokenflow}, ControlVideo \cite{zhang2023controlvideo}, SDEdit \cite{meng2022sdedit}, Stream Diffusion \cite{kodaira2023streamdiffusion}, and StreamV2V \cite{streamv2v}. We adopt a Two-Alternative Forced Choice (2AFC) protocol used in \cite{rerender, geyer2023tokenflow}, where participants are shown the input video, our video result, and a competing method's result. Users are then asked to determine which video has the better overall quality considering temporal consistency and alignment with the text prompt.

For each comparison, feedback was gathered from at least 12 different participants. Of note, our user study includes more participants than StreamV2V (3 participants) \cite{streamv2v}, FlowVid (5 participants) \cite{liang2023flowvid}, and ADA-VE (10 participants) \cite{mahmud2024ada}. The preference rates of PipeFlow versus the competing methods are shown in Table \ref{tab:User_Preference}.

\begin{table}[h]
\def\arraystretch{1.0}
\footnotesize
\setlength{\tabcolsep}{2pt}
\caption{\textbf{User preference study.} The rate of our PipeFlow being preferred is denoted as "Ours Preferred". The rate of the competing method being preferred is denoted as "Theirs Preferred". As shown, our approach is preferred over competing methods.}
\centering
\begin{tabular}[t]{|c|c|c|c|c|c|c|c|}
\hline
\textbf{Method} & \textbf{Ours Preferred (\%)} & \textbf{Theirs Preferred (\%)} \\ \hline
ControlVideo \cite{zhang2023controlvideo}  &  \textbf{84.1}   &  15.9   \\ \hline
TokenFlow \cite{geyer2023tokenflow} &  \textbf{55.8}   &  44.2   \\ \hline
SDEdit \cite{meng2022sdedit} & \textbf{70.8}  &  29.2  \\ \hline
StreamDiffusion \cite{kodaira2023streamdiffusion} & \textbf{96.6} &  3.4 \\ \hline
StreamV2V \cite{streamv2v} & \textbf{85.8}     &  14.2 \\ \hline
\end{tabular}
\label{tab:User_Preference}
\end{table}

We can see that PipeFlow is strongly preferred over methods such as ControlVideo \cite{zhang2023controlvideo}, StreamV2V \cite{streamv2v}, and StreamDiffusion \cite{kodaira2023streamdiffusion}. PipeFlow is preferred over TokenFlow \cite{geyer2023tokenflow} as well with a rate of 55.8\% preference over TokenFlow. PipeFlow is also preferred over SDEdit \cite{meng2022sdedit} with a preference rate of 70.8\% for PipeFlow to SDEdit's \cite{meng2022sdedit} 29.2\% preference rate.

\section{Additional Ablation Studies}\label{suppl_sec:ablations}
We present sample qualitative visualizations of various densely and sparsely sampled keyframes used for guiding the joint editing process, as well as an ablation on the effects of not using interpolation to generate the low-motion frames.

\subsection{Ablation on Dense and Sparsely Sampled Keyframes}

The number of keyframes used to guide the joint editing process can be customized by the user. We provide an ablation study of two possible configurations $Sparse$ indicating 1 keyframe for every 10 frames and $Dense$ indicating 5 keyframes for every 10 frames. As shown in Figure \ref{fig:supple_sparse_dense_ablation_video_edits} densely sampled keyframes lead to slightly better visual quality. For example, the eyes of the panther can better be seen in the $Dense$ keyframes setting.

However, The downside of the $Dense$ keyframes setting is it can reduce editing speed by up to 2$\times$. Since there is a tradeoff between the $Sparse$ and $Dense$ keyframes settings, users can adjust this parameter based on their preference for speed versus edit quality.

\subsection{Ablation on Interpolation of Low Motion Frames}
\begin{figure*}[ht]
\centering
\includegraphics[width=\linewidth]{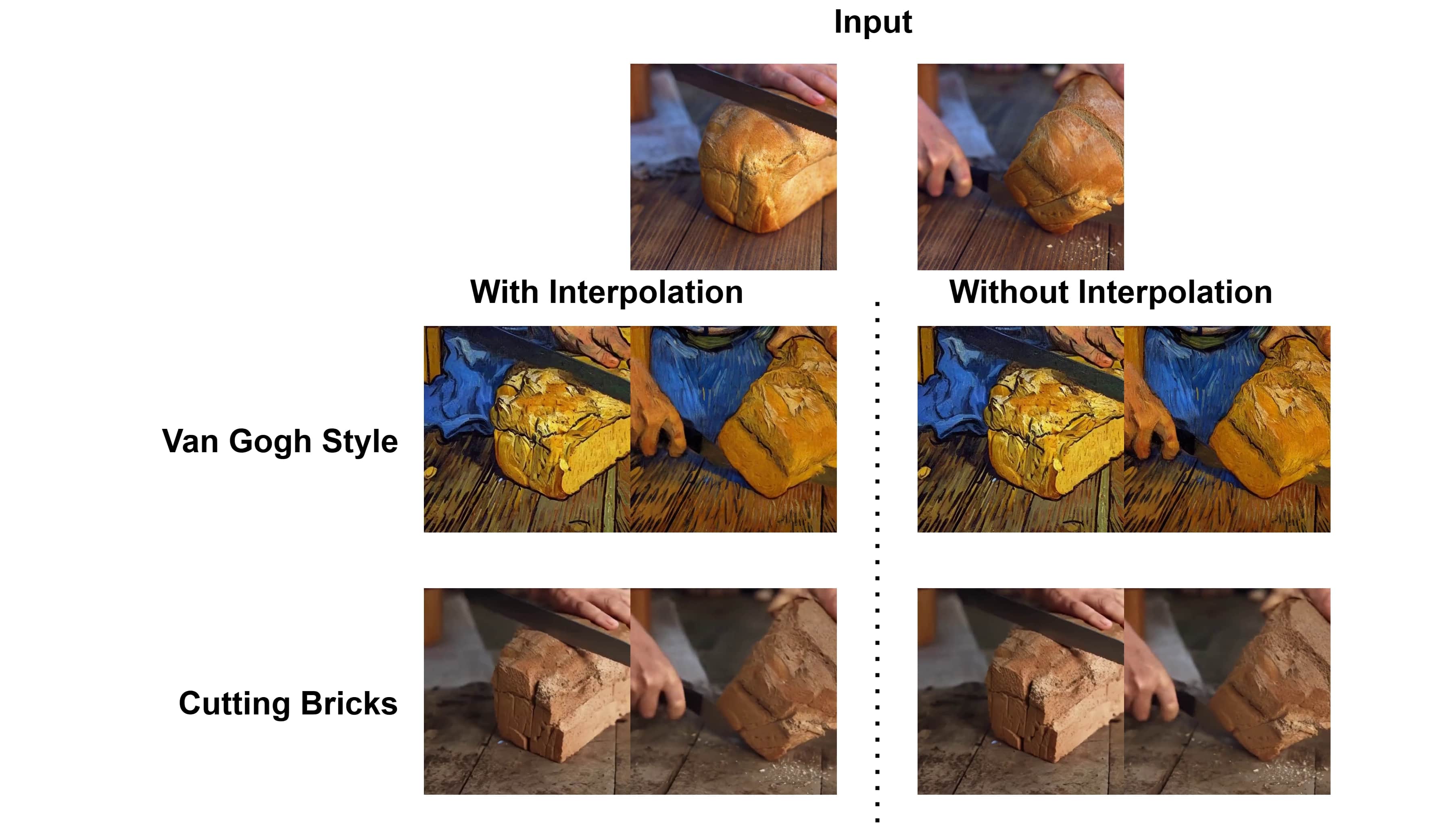}
\caption{\textbf{Ablation on Interpolation of Frames.} Whether performing style edits (e.g., Van Gogh style) or object edits (e.g., cutting bricks), interpolated frames are nearly identical to the original frames without interpolation.}
\label{fig:supple_ablation_video_edits}
\end{figure*}

As shown in Figure \ref{fig:supple_ablation_video_edits}, RIFE \cite{huang2022rife} interpolation effectively reconstructs skipped low-motion frames for editing, producing results with imperceptible differences to the human eye. This further validates our use of interpolation for skipping low-motion frames during editing. For instance, when editing the bread-cutting input to a Van Gogh style or transforming it to cutting bricks, the interpolated frames generated by RIFE \cite{huang2022rife} are nearly indistinguishable from the original frames without interpolation.

\subsection{Ablation on Increasing Border Frames Overlap to Improve Consistency}

\begin{figure*}[h]
\centering
\includegraphics[width=\linewidth]{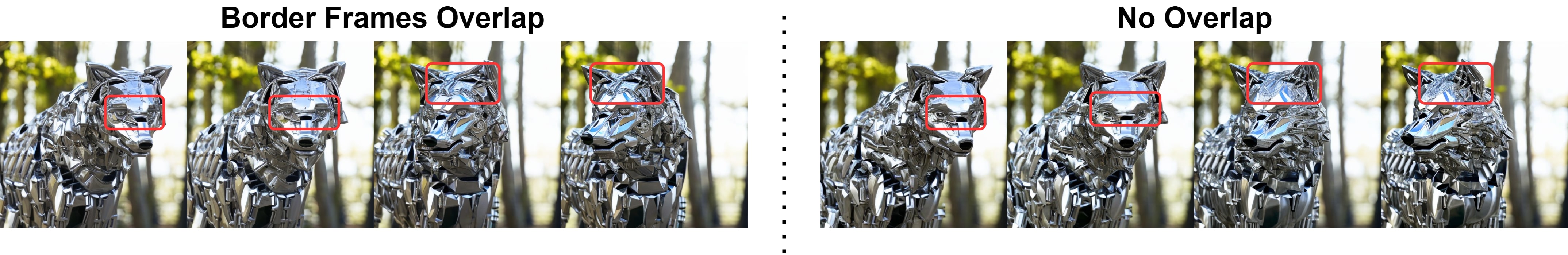}
\caption{\textbf{Ablation on Border Frames Overlap.} Consistency improves by adding border frames in PipeFlow. We use 5 overlapped border frames in the example.}
\label{fig:Rebuttal_Overlap_Frames}
\end{figure*}

Our interpolation model reduces visual artifacts at segment borders. To further improve consistency, we can overlap border frames from the previous segment in the next segment’s cross-attention. This approach improves the edit quality, as shown in Figure \ref{fig:Rebuttal_Overlap_Frames}, particularly in the consistency of the eyes and forehead, but reduces our speedup from \textbf{6$\times$ to 4.8$\times$}.

\begin{figure*}[ht]
\centering
\includegraphics[width=\linewidth]{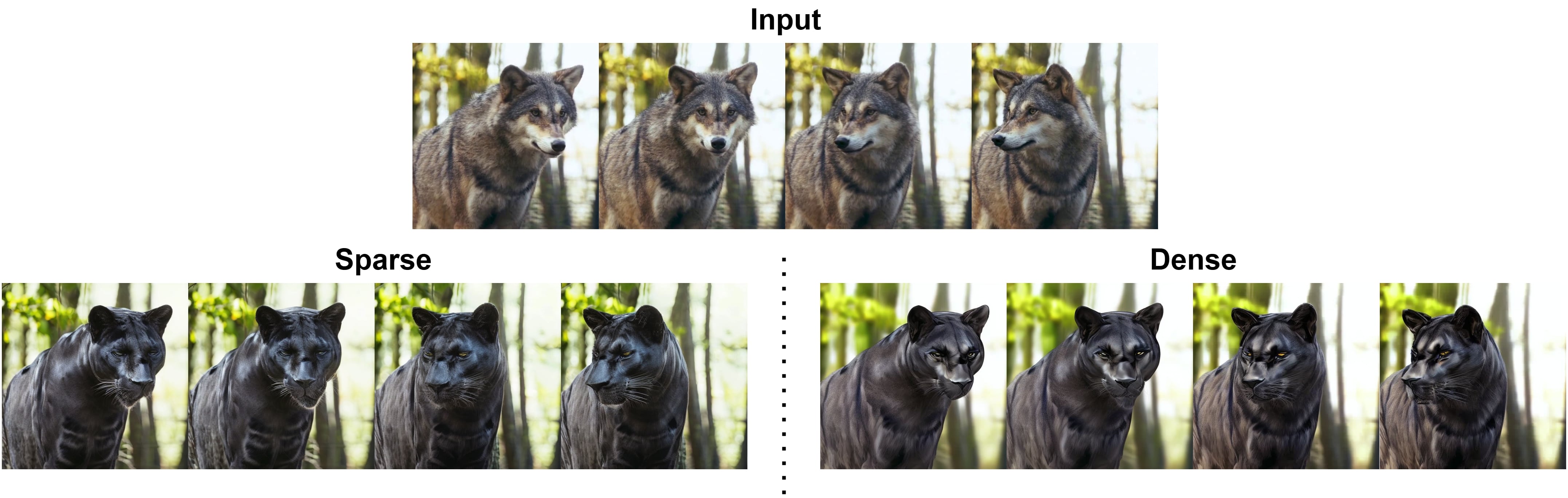}
\caption{\textbf{Ablation on Dense and Sparsely Sampled Keyframes.} Denser keyframes leads to slightly better edit quality and consistency as shown on the right. Sparser keyframes can lead to an up to 2$\times$ speedup compared to the dense setting.}
\label{fig:supple_sparse_dense_ablation_video_edits}
\end{figure*}

To verify the improvement, we assess the border frame consistency on 20 text-video prompt pairs by extracting motion vectors, warping the border frame and comparing the warped frame with the next frame via MSE and SSIM. Without overlap, we get an average MSE of 67.5 and an average SSIM of 0.564. With overlap, MSE (lower is better) improves to 66.8 (\textbf{0.7 improvement}), and SSIM (higher is better) improves to 0.580 (\textbf{0.016 improvement}). Using dense keyframes as shown in Figure \ref{fig:supple_sparse_dense_ablation_video_edits}, also reduces border frame inconsistencies.

\clearpage

\section{Qualitative Visualizations}\label{suppl_sec:visualizations}

\begin{figure*}[ht]
\centering
\includegraphics[width=\linewidth]{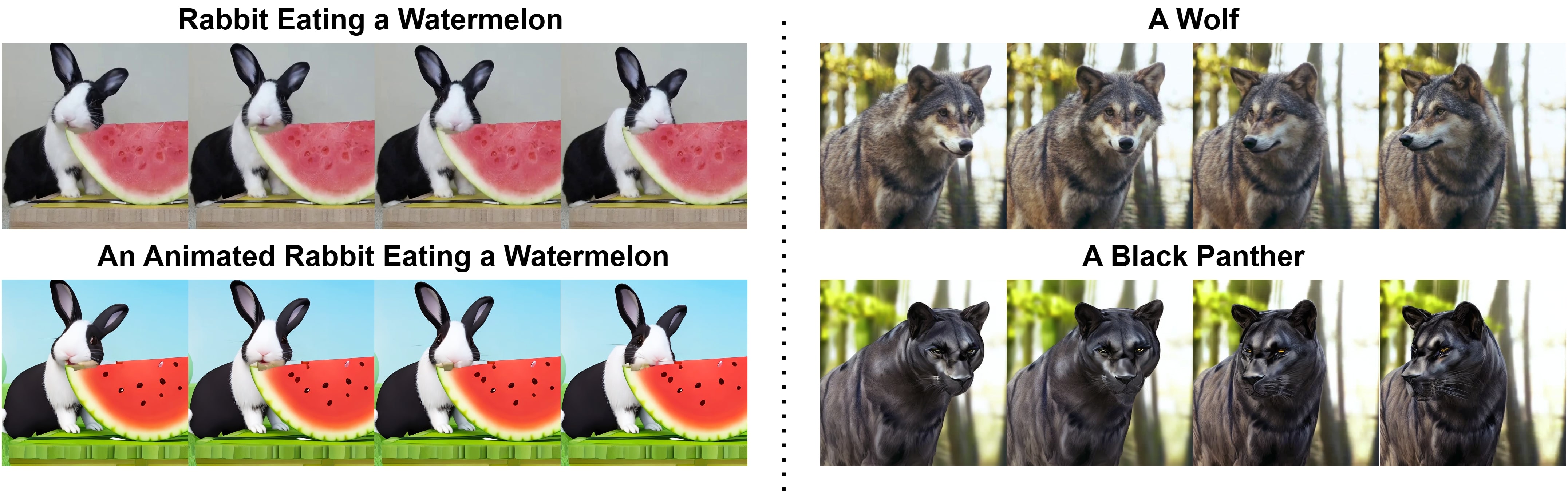}
\caption{\textbf{Additional Qualitative Visualizations of PipeFlow Edits.} PipeFlow demonstrates its capabilities in style transfer and object editing. In the first example, PipeFlow transforms the wolf into a black panther, preserving structure, shape, and color consistency. In the second example, PipeFlow applies a style edit to render the rabbit eating a watermelon in an animated aesthetic. The edits align with the provided prompts and maintain temporal consistency throughout the videos.}
\label{fig:supple_video_edits}
\end{figure*}

Qualitative examples of PipeFlow edits are shown in Figure \ref{fig:supple_video_edits}. In these examples, PipeFlow successfully modifies the wolf object into a black panther while maintaining structure, shape, and color consistency. Additionally, PipeFlow transforms the style of a rabbit eating a watermelon into an animated aesthetic, demonstrating the versatility of PipeFlow's possible edits.

\section{Limitations}\label{suppl_sec:limitations}

While PipeFlow demonstrates significant strengths, it also inherits some limitations from its underlying image editing method, Plug-and-Play Diffusion \cite{pnpDiffusion2023}. Specifically, PipeFlow struggles with major structural changes and fine-grained modifications, such as altering the color of an individual finger in the guidance video.

Additionally, PipeFlow inherits certain limitations from RIFE \cite{huang2022rife}, particularly in generating accurate interpolated images for low-motion frames skipped during diffusion-based editing.

Despite these challenges, the techniques introduced in this work are broadly applicable and can be integrated into most existing image and video editing frameworks.

\end{document}